\documentclass{article}

\usepackage{PRIMEarxiv}

\usepackage[utf8]{inputenc} 
\usepackage[T1]{fontenc}    
\usepackage{hyperref}       
\usepackage{url}            
\usepackage{booktabs}       
\usepackage{amsfonts}       
\usepackage{nicefrac}       
\usepackage{microtype}      
\usepackage{lipsum}
\usepackage{fancyhdr}       
\usepackage{graphicx}       
\graphicspath{{media/}}     

\usepackage{float}
\usepackage{placeins}

\title{A Contrastive Objective for Learning Disentangled Representations
}

\author{
  Jonathan Kahana, Yedid Hoshen \\
  School of Computer Science and Engineering \\
  The Hebrew University of Jerusalem, Israel \\
  \texttt{\{jonathan.kahana, yedid.hoshen\}@mail.huji.ac.il} \\
}

\begin{document}
\maketitle

\begin{abstract}

Learning representations of images that are invariant to sensitive or unwanted attributes is important for many tasks including bias removal and cross domain retrieval. Here, our objective is to learn representations that are invariant to the domain (sensitive attribute) for which labels are provided, while being informative over all other image attributes, which are unlabeled. We present a new approach, proposing a new domain-wise contrastive objective for ensuring invariant representations. This objective crucially restricts negative image pairs to be drawn from the same domain, which enforces domain invariance whereas the standard contrastive objective does not. This domain-wise objective is insufficient on its own as it suffers from shortcut solutions resulting in feature suppression. We overcome this issue by a combination of a reconstruction constraint, image augmentations and initialization with pre-trained weights. Our analysis shows that the choice of augmentations is important, and that a misguided choice of augmentations can harm the invariance and informativeness objectives. In an extensive evaluation, our method convincingly outperforms the state-of-the-art in terms of representation invariance, representation informativeness, and training speed \footnote{Code available at \href{https://github.com/jonkahana/DCoDR}{https://github.com/jonkahana/DCoDR}}. Furthermore, we find that in some cases our method can achieve excellent results even without the reconstruction constraint, leading to a much faster and resource efficient training.

\end{abstract}


\section{Introduction}
\label{sec:intro}

Representing the attributes of an image that are independent of its domain (e.g. imaging modality, geographic location, sensitive attribute or object identity) is key for many computer vision tasks. For instance, consider the following toy example: assume that we observe images of faces, each image is specified by the identity and pose but only labels of the identity are provided. The goal is to learn a representation that captures the unlabeled pose attribute, and carry no information about the identity attribute. This task has many other applications, including: learning to make fair decisions, cross domain matching, model anonymization, image translation etc. It is a part of the fundamental machine learning problem of representation disentanglement. We note that the most ambitious disentanglement setting, i.e. unsupervised disentanglement where no labels are provided, was proven by Locatello et al. \cite{locatello2019challenging} to be impossible without inductive biases. Luckily, our setting is easier than unsupervised disentanglement as the domain label is provided for all training images. This setting has attracted much research e.g. DRNET \cite{denton2017drnet}, ML-VAE \cite{bouchacourt2018mlvae} and LORD \cite{gabbay2020lord}. 

We begin by defining the desired properties for domain disentanglement. This task has two objectives: i) \emph{Invariance}: the learned representation should be invariant to the domain ii) \emph{Informativeness}: the learnt representation should include the information about all of the attributes which are independent of the domain. The invariance requirement is challenging, but it can \textit{in-principle} be directly optimized as the domain label is provided, e.g. using an adversarial discriminator. The informativeness requirement, however, is not generally possible to directly optimize without additional inductive biases as the attributes are unlabeled. This was theoretically demonstrated by \cite{Johansson2019SupportAI,invforda19zhao}. Nonetheless, recent methods have been able to achieve meaningful representations in many cases, by enforcing a reconstruction term, which optimizes a related objective.

\begin{figure}[t!]
    \centering
    \includegraphics[scale=0.33]{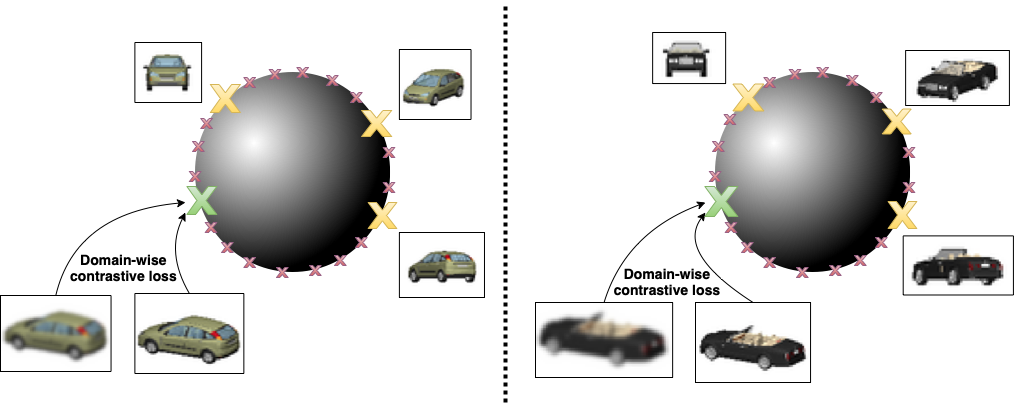}
    \caption{\textbf{\textit{An illustration of our method.}} The representations are domain invariant as the representations of each domain follow a spherically uniform distribution (encouraged by our domain-wise contrastive objective). Image augmentations (here Gaussian blurring) are used to assign similar images to nearby representations which indirectly improves informativeness. The reconstruction objective and encoder pre-trained weights initialization are not shown in this diagram.}
    \label{fig:model_diagram}
\end{figure}

We present a new method, \textbf{DCoDR}: \textbf{D}omain-wise \textbf{Co}ntrastive \textbf{D}isentangled \textbf{R}epresentations, that significantly improves both representation domain invariance and informativeness. To enforce the domain invariance, we propose a per-domain contrastive loss, that requires the representations of each domain to be uniformly distributed across the unit sphere. Differently from standard contrastive losses \cite{chen2020simclr}, our objective only considers negative examples from the \textit{same} domain. As shown in Sec. \ref{sec:ablations}, this seemingly simple change is crucial for learning domain invariant representations. Unfortunately, we find that encoders which satisfy this invariance constraint alone, are often uninformative over the desired attributes. This is a case of the documented phenomenon of \textit{feature suppression} \cite{featsupprunsup20li,intriguingcl20chen,avoidshort21robinson}. In line with previous methods \cite{gabbay2020lord,denton2017drnet,bouchacourt2018mlvae}, we optimize the informativeness of the representations indirectly by a reconstruction constraint. As we find this may be insufficient for learning informative representations in some cases, we propose two other techniques: i) Similarly to several self-supervised objectives (e.g. the one in SimCLR \cite{chen2020simclr}), we enforce representations of images to be similar to those of their augmentations. Despite being common among self-supervised methods, we show that standard choices of augmentations (specifically, those used by SimSiam \cite{chen2020simsiam}) can harm the domain invariance of the representation. We analyse the effectiveness of different augmentations for domain invariant representation learning. ii) Initializing the image encoder using weights pre-trained with self-supervision on an external dataset, which 
we empirically find to learn both more informative and invariant representations.

We evaluate our method on five popular benchmarks. Our method significantly exceeds the state-of-the-art in terms of invariance and informativeness. We investigate a fully discriminative version and find that in many cases it is competitive with the previous state-of-the-art while being much faster.

\noindent A summary of our contributions:
\begin{enumerate}
    \item A non-adversarial and non-generative, domain invariance objective.
    \item Analysing the benefits and pitfalls of image augmentations for informativeness and domain invariance of the learned representations.
    \item A new approach, DCoDR, which significantly outperforms the state-of-the-art in domain invariant representation learning.
    \item A discriminative only variant, which is 5X faster than existing approaches.
    \item An extensive evaluation on five datasets.
\end{enumerate}

\section{Related Work}
\label{related_works}

\textbf{Learning domain disentangled representations.} Much research was done on separating between labeled and unlabelled attributes. Several methods use adversarial training  \cite{denton2017unsupervised,szabo2017challenges,mathieu2016disentangling}. Other methods use non-adversarial approaches, e.g. cycle consistency \cite{harsh2018disentangling}, group accumulation \cite{bouchacourt2018multi} or latent optimization \cite{gabbay2020lord,gabbay2021scaling}. Our method improves upon this body of work. 

\textbf{Contrastive representation learning.}  Significant progress in self-supervised representation learning was achieved by methods relying on pairs of augmented samples. Most recent methods use the constraint that the neural representations of different augmentations of the same image should be equal. Non-contrastive methods \cite{chen2020simsiam,grill2020byol,richemond2020byolworks} use the above constraint with various other tricks for learning representations. As the above formulation is prone to collapse, contrastive methods \cite{ye2019spreading,hjelm2019mutual,wu2018discrimination,oord2018coding,hjelm2019mutual,he2020moco,chen2020mocov2,misra2020pirl,chen2020simclr,chen2020simclrv2} add an additional uniformity constraint that prohibits collapse of the representation to a single point. We propose a per-domain contrastive objective, tailored for domain disentanglement. 

\textbf{Contrastive approaches for disentanglement.} Recently, Zimmerman et al. \cite{pmlr-v139-zimmermann21a} proposed a seminal approach for  contrastive learning of disentangled representations. They tackle the ambitious setting of unsupervised disentanglement, and therefore make strong assumptions on the distribution of the true factors of variation as well as requiring temporal sequences of images at training time. Our method applies to the different (and less ambitious) setting of domain disentanglement - assuming domain labels for training data, but not having image sequences or strong assumptions on the evolution of unlabeled true factors. Our technical approaches are consequently very different. 

 \textbf{Applications of disentangled representations.} Learning disentangled representations has many applications including: controllable image generation \cite{zhu2018visual}, image manipulation \cite{gabbay2020lord,gabbay2021scaling,wu2021stylespace} and domain adaptation \cite{peng2019domain}. Furthermore, it is believed that better disentangled representations will have future impact on model interpretability \cite{hsu2017interpretable}, abstract reasoning \cite{van2019reasoning} and fairness \cite{creager2019flexiblyfair}.

\section{Domain Invariant Representation Learning}
\label{sec:alignment}

\subsection{Preliminaries}

We receive as input a set of training samples $\mathcal{X}_t = \{x_1,x_2,..,x_N\}$. Each training sample $x \in \mathcal{X}_t$ has a labeled domain $d$ and unlabelled attributes $y$ which are uncorrelated to $d$.  We assume that the labeled domain $d$ is a single categorical variable. The objective is to learn an encoder $E$, which encodes each image $x$ as code $z = E(x)$ satisfying the criteria in Sec.~\ref{sec:crit}. 

\subsection{Criteria} 
\label{sec:crit}

The domain disentanglement task requires satisfying the following two criteria:

\textbf{Invariance}: We require that the representation $z$ should not be predictive of the domain $d$. This can be written as:
\begin{equation}
\label{eq:disentanglement_metric}
    P(d|z) = P(d)
\end{equation}

\textbf{Informativeness}: We require that the representation $z$ should encapsulate as much information on attributes $y$ as possible. Note that $z$ cannot hold more information about $y$ than the original image $x$, as there exists a deterministic encoder $E$ which maps $x$ to $z$. It therefore follows by the data processing inequality, that the maximally informative representation $z$ should be as informative as the original image about the attributes $y$: 
\begin{equation}
\label{eq:completeness_metric}
    I(y,z) = I(y,x)
\end{equation}




In our setting, only the domain labels $d$ are provided but not the attribute labels of $y$. The objective in Eq.~\ref{eq:completeness_metric} cannot therefore be optimized directly. Saying that, in line with previous methods, we optimize informativeness by training a conditional generator through a reconstruction objective. Unlike previous methods, we use additional techniques which increase informativeness significantly.
Our proposed approach will be detailed in Sec.~\ref{sec:augmentations}.

\subsection{Existing Approaches for Invariance Optimization}
\label{subsec:prev_inv}

Current methods optimize the invariance criterion using two main approaches: 

\textbf{Adversarial methods \cite{denton2017drnet}.} Many disentanglement methods rely on adversarial domain confusion constraints to ensure representation invariance. They are often written in the following form:

\begin{equation}
\label{eq:invariance_gan}
    L_{adv} = \max_D \ell_{CE}(D(E(x)), d)
\end{equation}

\noindent Where $\ell_{CE}$ is the cross-entropy loss. The discriminator $D$ measures how informative the representation $z=E(x)$ is over the original domain $d$. An encoder that satisfies this constraint will indeed be domain invariant $P(d|z) = P(d)$. Unfortunately, adversarial training is challenging and the optimization often fails to minimize this loss perfectly.

\textbf{Variational-autoencoders (VAE) \cite{bouchacourt2018mlvae,gabbay2020lord}.} Given the weaknesses of adversarial methods, variational methods were proposed that ensure the representations are normally distributed  $P(z|d) = N(0; I)$. The encoder in this case outputs the parameters of a Gaussian distribution of the posterior $p(z|x)$. Using the ELBO criterion, the objective becomes:
\begin{equation}
\label{eq:invariance_VAE}
    L_{vae} = \ell_{KL}(E(x), N(0, I))
\end{equation}

\noindent However, LORD \cite{gabbay2020lord} found that simply optimizing this criterion does not converge to disentangled representations. Furthermore, they showed that randomly initialized encoders are highly entangled and variational losses were insufficient for removing this entanglement. Instead, they suggested using latent optimization rather than deep encoders at first, for directly learning the representation $z$ of each training image $x$. This indeed improves the domain invariance of the representations, but is more sensitive to hyper-parameter choices. It also requires an inconvenient and time consuming second stage, for learning an image to representation encoder.

\section{DCoDR: Learning Domain-wise Contrastive Disentangled Representation}
\label{sec:method}

\subsection{Overview}

We introduce a new approach, DCoDR, for learning informative, domain invariant representations. In Sec.~\ref{sec:invariance_objective}, a new per-domain contrastive loss is proposed to enforce invariance directly. It does not, by itself, require the representation to be maximally informative. To overcome this issue, we optimize informativeness indirectly by \textit{reconstruction} and \textit{image augmentation} objectives as well as \textit{encoder pre-trained weights initialization}. We investigate an additional, fully discriminative variant of our method, which is much faster than existing methods at the price of lower informativeness.

\subsection{Representation Invariance with Domain-wise Contrastive Losses}
\label{sec:invariance_objective}

Learning an invariant representation requires the domain $d$ to be unpredictable from the learned representation $z$. We present a non-adversarial method for encouraging domain invariance.
Our approach enforces the probability distribution of representations $z$ to follow a uniform spherical distribution (denoted $U_S$) regardless of the domain $d$: $P(z|d) = U_S$. It follows from Bayes' law that the representation $z$ does not provide any information about the domain, $\forall z: P(d|z) = P(d)$. This also yields that mutual information between the domain and representation is zero $I(d, z) = 0$.

The above analysis requires that $P(z|d) = U_S$ for every domain $d$. We do so by training a separate contrastive loss for every domain $d$. It was highlighted by Wang and Isola \cite{wang2020understanding} that the denominator of the contrastive objective encourages the representations follow a uniform spherical distribution. Learning a contrastive loss \textit{separately} over image representations from different domains, ensures that the representations $z$ are distributed as $U_S$ regardless of the domain $d$. For an image $x$ from domain $d$, this can be written as follows:

\begin{equation}
\label{eq:contrastive_objective}
    \mathcal{L}_{inv}(x,d) = \log{\sum_{(x', d') \in \mathbb{X}} \textbf{1}_{d' = d} e^{sim(E(x'),E(x))}}
\end{equation}

\noindent $sim$ is a similarity function, cosine similarity in our case. The objective only considers image pairs drawn from the \textit{same} domain. Unlike previous methods in Sec.~\ref{subsec:prev_inv}  
(e.g. \cite{denton2017drnet,moyer2018invariant,gabbay2020lord}),
it does not rely on adversarial or variational approximations.

\subsection{Improving Representation Informativeness}
\label{sec:augmentations}

Beyond invariance, the representations $z$ should encapsulate the information about all of the image attributes $y$ except the domain label $d$. In Eq.~\ref{eq:completeness_metric} this was shown to imply $I(y,x)=I(y,z)$. 
We cannot directly optimize this constraint, as the attributes $y$ for image $x$ are not provided in our setting. In line with previous methods presented in Sec.~\ref{subsec:prev_inv}, we optimize the informativeness indirectly by a reconstruction constraint. Furthermore, we present two algorithmic choices that empirically further increase informativeness significantly.

\textbf{Reconstruction:} Reconstruction constraints are an established way to improve the informativeness of the representation $d$. They have been used in many previous methods \cite{denton2017drnet,bouchacourt2018mlvae,gabbay2020lord}. In line with previous methods, we include a reconstruction constraint in our method. Specifically, we learn a conditional generator $G$ that takes as input the domain $d$ and representation $z$ and outputs an image $G_d(z)$. The reconstruction objective requires that the output image is as close as possible to the input image $x$. The difference between the reconstruction and original images is measured using the function $\ell$. In practice, we use the same perceptual loss as in LORD \cite{gabbay2020lord} in Eq.~\ref{eq:rec_gen}.

\begin{equation}
\label{eq:rec_gen}
    L_{rec} = \sum_{d \in {\cal D}}\sum_{x \in {\cal X}_d} \ell_{perc}(G_d(E(x)),x)
\end{equation}

\textbf{Augmentations:} Contrastive objectives are susceptible to shortcut solutions that lower informativeness, also known as \textit{feature suppression} \cite{avoidshort21robinson}. This occurs by (inadvertently) learning an encoder that maps nuisance image attributes (or noise) to the spherical uniform distribution. This representation ignores the other image attributes, therefore being insufficiently informative. Ensuring that image augmentations have similar representations to the original image can help reduce this collapse, for suitably well selected augmentations:

\begin{equation}
\label{eq:aug_objective}
    \mathcal{L}_{aug}(x) = -sim(E(A_1(x)),E(A_2(x))
\end{equation}

\noindent Where $A_1(x)$ and $A_2(x)$ are two random augmentations of image $x$. Unfortunately, poorly selected augmentations can make the representation $z$ invariant to the desired attributes $y$, which is harmful. E.g. when $y$ is pose, and the augmentation is horizontal flipping, the representation $z$ will be invariant to flip direction, therefore less informative over the pose. To test this hypothesis, we trained our method's discriminative variant, DCoDR-norec, using blur or flip augmentations on Cars3D. We measure each metric as explained in Sec.~\ref{subsec:eval_rep}. Tab. \ref{tab:aug_selection} shows flipping significantly reduced the informativeness. 

\begin{table}[h!]
    \centering
    \caption{\textbf{\textit{DCoDR-norec with Different Augmentations.}} Evaluation of our method's discriminative variant, DCoDR-norec, with $2$ different augmentations on Cars3D. 
    \label{tab:aug_selection}
    }
        \begin{tabular}{lcc}
             & Inv. $(\downarrow)$ & Inform. $(\uparrow)$ \\
             \midrule
            Gaussian Blurring & 0.002 & 0.960 \\
            Horizontal Flipping & 0.003 & 0.725 \\
            \bottomrule
    \end{tabular}
\end{table}

It is clear from the discussion above that augmentations can be highly desirable for improving informativeness, while their choice is important. We discovered that the standard augmentations used by state-of-the-art contrastive methods e.g. \cite{chen2020simclr,grill2020byol,chen2020simsiam} are not optimal for our task. The reason is that they are designed to keep information only about the object's 'class' while being invariant to all other attributes. This may, in some cases, also require invariance on the attributes of interest $y$. Instead, we selected a much smaller set of augmentations which we empirically show to be effective on a set of datasets that we considered. The selected augmentations are: i) Cropping, ii) Gaussian Blurring, iii) Increase of contrast iv) Increase in saturation. For Edges2Shoes \cite{yu2014edges2shoes} dataset, we find it more effective to include gaussian blurring alone. In Sec.~\ref{sec:ablations} we show the selected augmentations significantly outperform the standard set of SimSiam \cite{chen2020simsiam}.

\textbf{Encoder Initialization with Unsupervised Pre-Trained Weights:} Although the constraints proposed in this section are effective for learning domain disentangled representations, we empirically find they are not always sufficient.
In order to improve generalization \cite{erhan10a}, we propose to initialize the encoder with the weights of a network pre-trained in an \textbf{unsupervised} manner (MoCo-V2 \cite{chen2020mocov2}) on the ImageNet dataset. Using the inductive bias from pre-trained weights in this setting is common, e.g. LORD \cite{gabbay2020lord} uses an ImageNet pre-trained perceptual loss.
Note, this initialization is not beneficial for LORD as it does not use an encoder in the first stage.

\subsection{Our Complete Method: DCoDR}

DCoDR optimizes the combination of the $3$ objectives presented in this section:

\begin{equation}
        \label{eq:total_objective}
            \min_{E,G} L_{DCoDR} = L_{inv} + L_{rec} + L_{aug}
\end{equation}

\noindent We use the augmentations from Sec.~\ref{sec:augmentations}. We initialize the encoder $E$ with the weights of an MoCo-V2 encoder pre-trained on ImageNet (without labels).

\subsubsection{Discriminative DCoDR (DCoDR-norec)} We present a discriminative variant of our method, by simply dropping the reconstruction constraint:

\begin{equation}
        \label{eq:abcd_total_objective}
            \min_{E} L_{DCoDR-norec} = L_{inv} + L_{aug}
\end{equation}

\noindent The lack of a reconstruction constraint, makes this variant typically learn less informative representations than DCoDR. However, as this variant does not train a generator, it is several times faster than DCoDR which by itself is considerably faster than previous state-of-the-art LORD.

\subsection{Differences From SimCLR}
\label{sec:simclr}

Although a part of our method is motivated by the SimCLR \cite{chen2020simclr} objective, it is significantly different. In Tab. \ref{tab:ablations} we show that although the differences from SimCLR might look superficially simple, each of them is essential for the success of our method (the first $3$ apply for DCoDR-norec as well):
\begin{itemize}
    \item \textbf{Domain-wise Loss.} DCoDR learns a contrastive loss over each domain separately whereas SimCLR learns a single loss over all the data.
    \item \textbf{Choice of Augmentations.} DCoDR learns a reduced set of augmentations rather than the standard set used in SimCLR.
    \item \textbf{Pre-Training.} DCoDR initializes the encoders weights by \textit{unsupervised} pre-training on ImageNet using MoCo-V2 \cite{chen2020mocov2}, which does \emph{not} use any labels.
    \item \textbf{Reconstruction} DCoDR uses a reconstruction term for increasing the informativeness of its representations, which does not exist in SimCLR.
\end{itemize}

\section{Experiments}
\label{sec:exp}

In this section, we evaluate our method against (variational and adversarial) state-of-the-art domain disentanglement approaches. We evaluate the invariance and informativeness of the learned representations. We then demonstrate cross domain retrieval of our method compared to the other baselines in Sec. \ref{sec:retrieval}.

\textbf{Benchmark Datasets.} We report results on Cars3D \cite{carsdataset}, SmallNorb \cite{smallNORB}, Shapes3D \cite{3dshapes18}, CelebA \cite{CelebAMask-HQ} and Edges2Shoes \cite{yu2014edges2shoes}. All datasets are used in 64x64 resolution. Due to the large number of samples in the full Shapes3D and the limited variation between them, we randomly sample $50,000$ images for training, while keeping the test set size at $10\%$ of the original size. 

\subsection{Implementation Details}
\label{imp}

\textbf{Architecture and optimization.} We used a ResNet50 encoder, trained for $200$  epochs using a batch size of $128$. Each batch was composed from $32$ images drawn from $4$ different classes. In line with other methods e.g. LORD, the reconstruction loss is computed using a VGG based perceptual loss pre-trained on ImageNet.  

\textbf{Baselines.} We use the default parameters of ML-VAE \cite{bouchacourt2018mlvae} and DRNET \cite{denton2017drnet}. We tried to replace their encoders by larger ResNet architectures but this resulted in degraded performance. We therefore kept the original architectures and hyper-parameters for all runs. We use a ResNet50 architecture for LORD's second stage amortized encoder and train it for $200$ epochs. We do not compare to OverLORD \cite{gabbay2021scaling} as in our evaluated datasets it is exactly the same as LORD.

\textbf{Augmentations.} As mentioned in Sec. \ref{sec:augmentations}, we used cropping, Gaussian blurring, high contrast and high saturation transformations as our positive augmentations, except for Edges2Shoes where we use only Gaussian blurring. 

\subsection{Representation Evaluation}
\label{subsec:eval_rep}

\textbf{Experimental Setup.} For each dataset, we evaluate both \emph{Invariance} and \emph{Informativeness} of the representations. To do so, we train a deep classifier to predict all image attributes from the learned representations, including the domain $d$ and the other factors $y$. For the synthetic datasets, we compute each of the two objectives over each factor separately. Since some of the datasets have multiple factors, we present the average of the informativeness over all factors, while the full results are presented in App.~\ref{appendix:sec:complete_results}. For CelebA we use the location of the $68$ landmarks \cite{bulat2017far} as the uncorrelated attribute. As the landmarks are numeric rather than categorical, we train an MLP regression model to predict the landmark locations. We measure the $L_1$ error of the MLP regressor where lower errors are better. 
To understand how far the results are from the theoretical limit, we present the frequency of the most common domain value as a lower bound on the invariance. Note that since we use a probabilistic estimator to evaluate our metrics, in some cases (especially when performance is close to optimal limit) the invariance may be slightly lower than the theoretical limit. This can happen when the classifier slightly overfits its training data, hence the small gap.

\textbf{Results.} The results on all datasets are presented in Tab.~\ref{tab:disent_and_rep_quality}. We observe that on Cars3D, even though LORD is a strong baseline, both discriminative and complete variants of DCoDR are able to surpass it, and achieve nearly perfect results. ML-VAE and DRNET did not perform as well on this dataset, inline with the observations in \cite{gabbay2020lord}. On SmallNorb, it is clear that LORD fails to disentangle the domain. Both our methods outperforms it on both metrics, achieving much more disentangled representations than any other method. As the representations learned by ML-VAE and DRNET are not domain invariant, they have higher informativeness but do not satisfy the main requirement of disentanglement. Note that we used the original version of the SmallNorb benchmark rather than the simplified version presented in the LORD paper. In this setting, the domain is defined as the object category alone whereas \textit{both} pose and lighting are unknown. On Shapes3D, again both variants of DCoDR achieve almost perfect results while all other methods suffer from lack of domain invariance. LORD achieves very limited invariance while ML-VAE and DRNET learn representations that are not invariant at all. CelebA is challenging for our per-domain contrastive loss, as it contains very few images per each domain, meaning the estimation of a uniform distribution for each domain is limited. That being said, we observe DCoDR performs better than LORD. It has an additional advantage over LORD of not requiring 2-stage optimization. CelebA is a failure case for our discriminative variant. Although presenting stronger invariance than the other methods, it is not sufficiently informative about the landmarks. 

Generally, DCoDR demonstrated state-of-the-art results in invariance and informativeness. In some cases (e.g. CelebA), DCoDR-norec fails to learn sufficiently informative representations, while being more invariant than previous methods as well as DCoDR itself. We emphasize that a key advantage of DCoDR-norec is its training time. DCoDR-norec training is $5$ to $10$ times faster compared to LORD and $3$ times faster to compared to DCoDR (see Sec.~\ref{sec:run_time}). 

\begin{table}[h!]
    \centering
        \caption{\textbf{\textit{Representation Evaluation Results.}} Content Invariance $(\downarrow)$ (Content to Domain) and Representation Quality $(\uparrow)$ (Average Prediction Accuracy). For CelebA we use extracted landmarks as attributes, and compute the regression L1 $(\downarrow)$ error.}
        \label{tab:disent_and_rep_quality}
        \begin{tabular}{lcccccccc}
             & \multicolumn{2}{c}{\textbf{Cars3D}} & \multicolumn{2}{c}{\textbf{SmallNorb}} & \multicolumn{2}{c}{\textbf{Shapes3D}} & \multicolumn{2}{c}{\textbf{CelebA}} \\
             \cmidrule(lr){2-3} \cmidrule(lr){4-5} \cmidrule(lr){6-7} \cmidrule(lr){8-9} & Inv. $(\downarrow)$ & Inform. $(\uparrow)$ & Inv. $(\downarrow)$ & Inform. $(\uparrow)$ & Inv. $(\downarrow)$ & Inform. $(\uparrow)$ & Inv. $(\downarrow)$ & L1 $(\downarrow)$ \\
             \midrule
            LORD & 0.009 & 0.940 & 0.393 & 0.670 & 0.703 & 0.995 & 0.019 & 0.862 \\
            DRNET & 0.504 & 0.909 & 0.953 & 0.899 & 0.892 & 1 & 0.084 & 0.795 \\
            ML-VAE & 0.697 & 0.930 & 0.968 & 0.944 & 0.999 & 1 & 0.136 & 0.723 \\
            \midrule
            DCoDR-norec & 0.005 & 0.970 & 0.071 & 0.730 & 0.246 & 0.997 & 0.015 & 1.127 \\
            DCoDR & 0.005 & 0.980 & 0.143 & 0.785 & 0.245 & 0.999 & 0.017 & 0.858 \\
            \midrule
            Optimal & 0.005 & 1 & 0.021 & 1 & 0.251 & 1 & 0.002 & 0 \\
            \bottomrule
    \end{tabular}
\end{table}

\subsubsection{Ablation Study} 
\label{sec:ablations}
We ablate our method on the SmallNorb dataset (Tab. \ref{tab:ablations}). First, we observe that removal of the unsupervised MoCo-V2 pre-trained weight initialization significantly hurts all metrics. Removal of per-domain negative pairs i.e. using a single contrastive loss for all domains (the loss used in SimCLR), makes the representations entangled. We also tested removing the positive augmentations, using the objective in Eq.~\ref{eq:contrastive_objective}. Removing the positive augmentations has different effects in to DCoDR and DCoDR-norec. DCoDR's informativeness was reduced while invariance improved. DCoDR-norec fails without the positive augmentations as they are its only objective that enforces informativeness. Lastly, we consider the standard set of augmentations used in SimSiam \cite{chen2020simsiam}. This choice significantly harms both invariance and informativeness in both variants.

\begin{table}[h!]
    \centering
        \caption{\textbf{\textit{Ablation Analysis of Our Method.}} Replacing the per-domain contrastive loss by a standard one severely hurts invariance. Augmentations are crucial for the DCoDR-norec. Pre-training improves both metrics, mostly invariance. The selection of augmentations is important and can harm the invariance.
        }
        \label{tab:ablations}
        \begin{tabular}{lcccc}
            & \multicolumn{2}{c}{\textbf{DCoDR}} & \multicolumn{2}{c}{\textbf{DCoDR-norec}} \\
            \cmidrule(lr){2-3} \cmidrule(lr){4-5}
            & Inv. $(\downarrow)$ & Inform. $(\uparrow)$ & Inv. $(\downarrow)$ & Inform. $(\uparrow)$ \\ 
            \midrule
            No Domain Negatives & 0.863 & 0.829 & 0.879 & 0.754 \\
            No Positive Augmentations & 0.057 & 0.555 & 0.021 & 0.166 \\
            No Pre-Training & 0.253 & 0.701 & 0.298 & 0.716 \\
            SimSiam \cite{chen2020simsiam} Augmentations & 0.244 & 0.643 & 0.246 & 0.658 \\
            Complete Method & 0.143 & 0.785 & 0.071 & 0.730 \\
            \midrule
            Optimal & 0.020 & 1 & 0.020 & 1 \\
            \bottomrule
    \end{tabular}
\end{table}

\subsection{Cross Domain Retrieval Evaluation}
\label{sec:retrieval}
\textbf{Experimental Setup.} We evaluate the representations $z$ learned by our method against the baselines on cross domain retrieval. For each image $x$ in the test set, we first extract its representation $z=E(x)$ and retrieve its nearest neighbors using $L_2$ distance. We present both quantitative and qualitative analyses. For our quantitative analysis, we use the labels of the attributes $y$ for deciding whether a match was found or not. Since many attributes are naturally ordered we would like to consider more than just perfect matches in all attributes. To do so, we allow a match for small changes in some numeric attributes, as detailed in App.~\ref{appendix:sec:complete_results}. Here we present the accuracy of matching over all attributes. The accuracy of matching individual attributes is presented in App.~\ref{appendix:sec:complete_results}. We also visually present the $5$ nearest-neighbor images for several test set images - using the representations learned by our and baseline methods. The analysis can highlight leakage of domain information in learned representations, as will be shown below.

\begin{figure}[t!]
\centering
\begin{tabular}{c@{\hskip8pt}|@{\hskip8pt}c}
DRNET & ML-VAE \\ 
\includegraphics[width=0.4\linewidth]{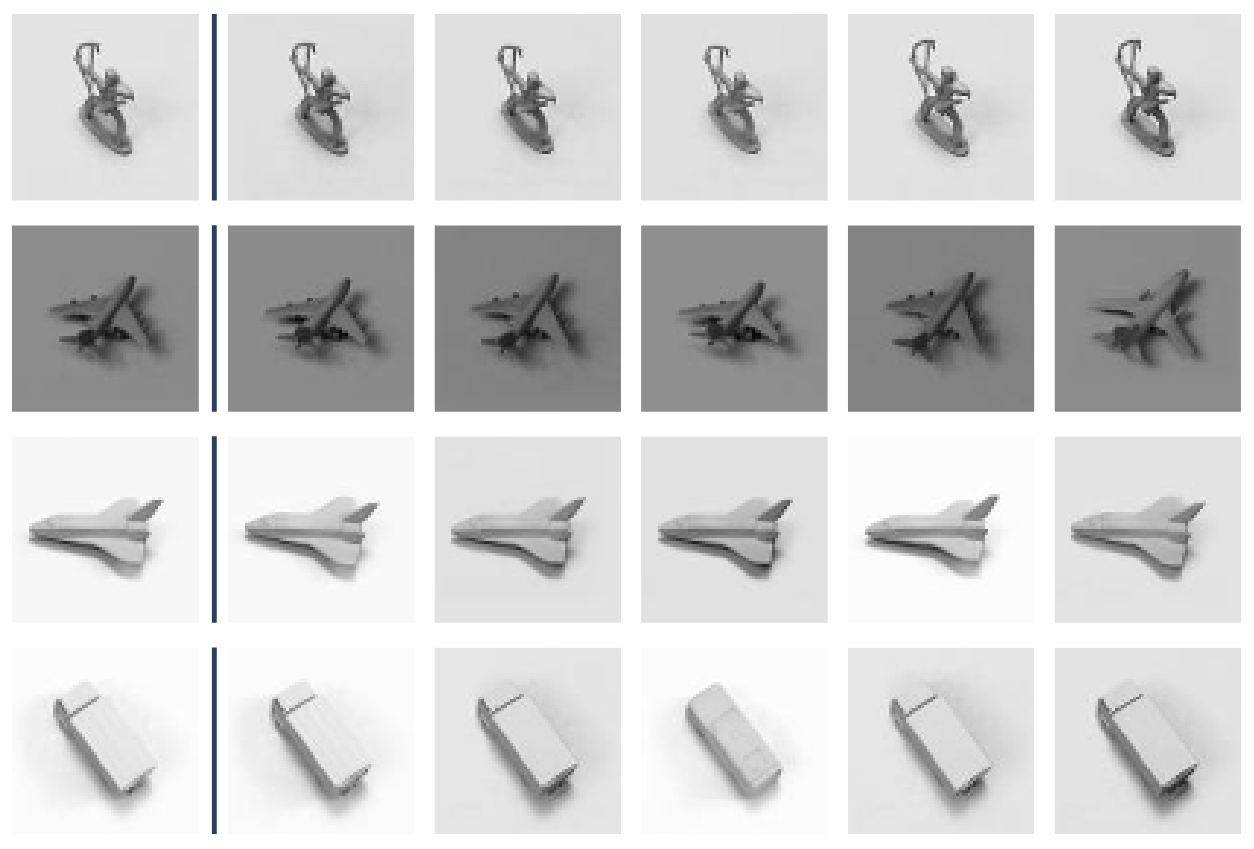} & 
\includegraphics[width=0.4\linewidth]{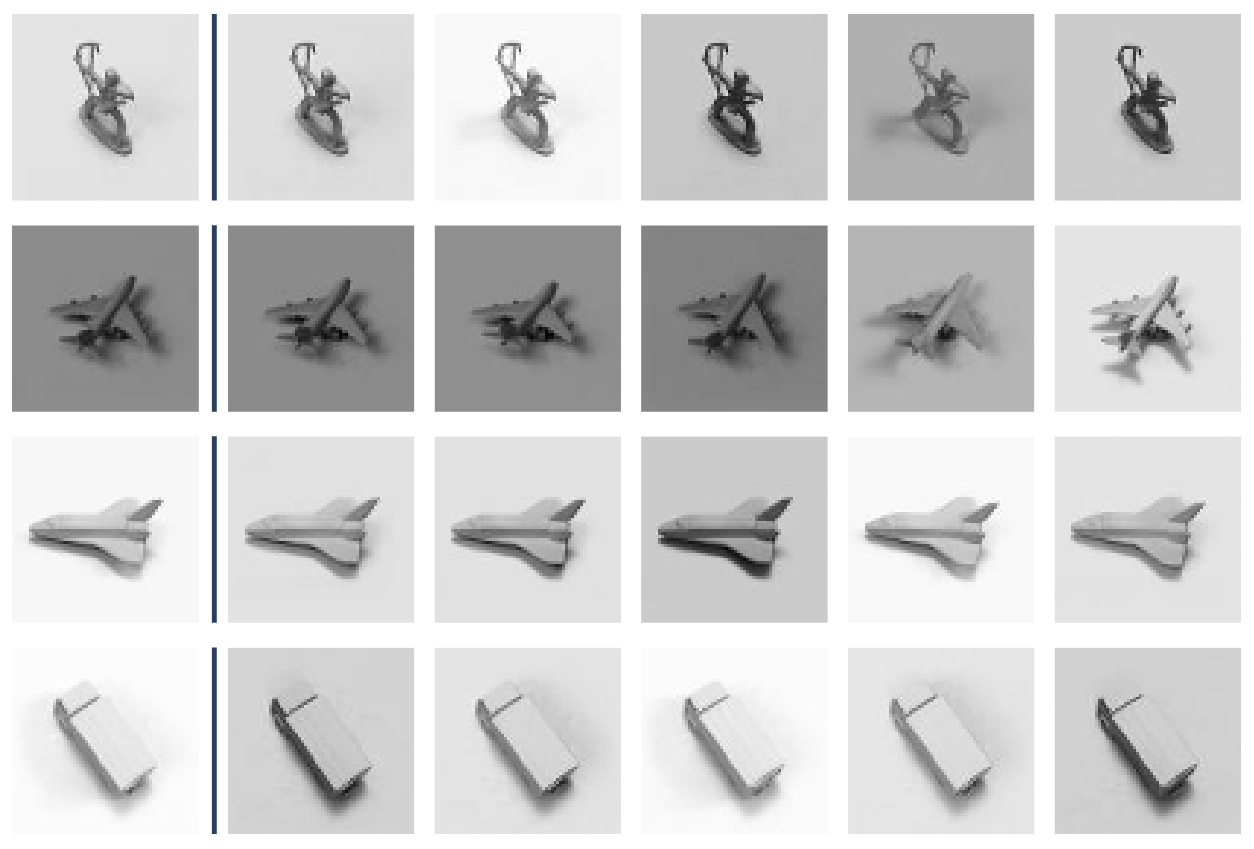} \\
\midrule
LORD & DCoDR-norec  \\
\includegraphics[width=0.4\linewidth]{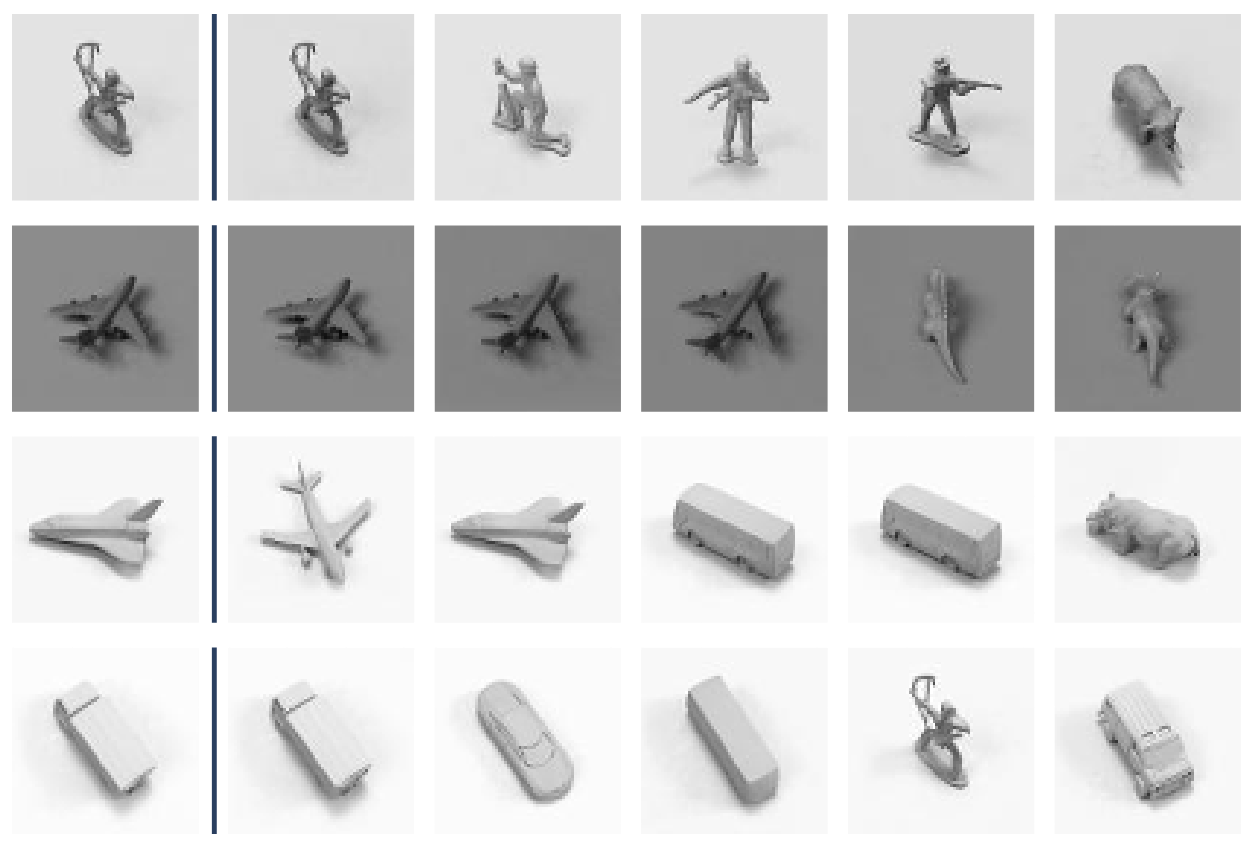} & 
\includegraphics[width=0.4\linewidth]{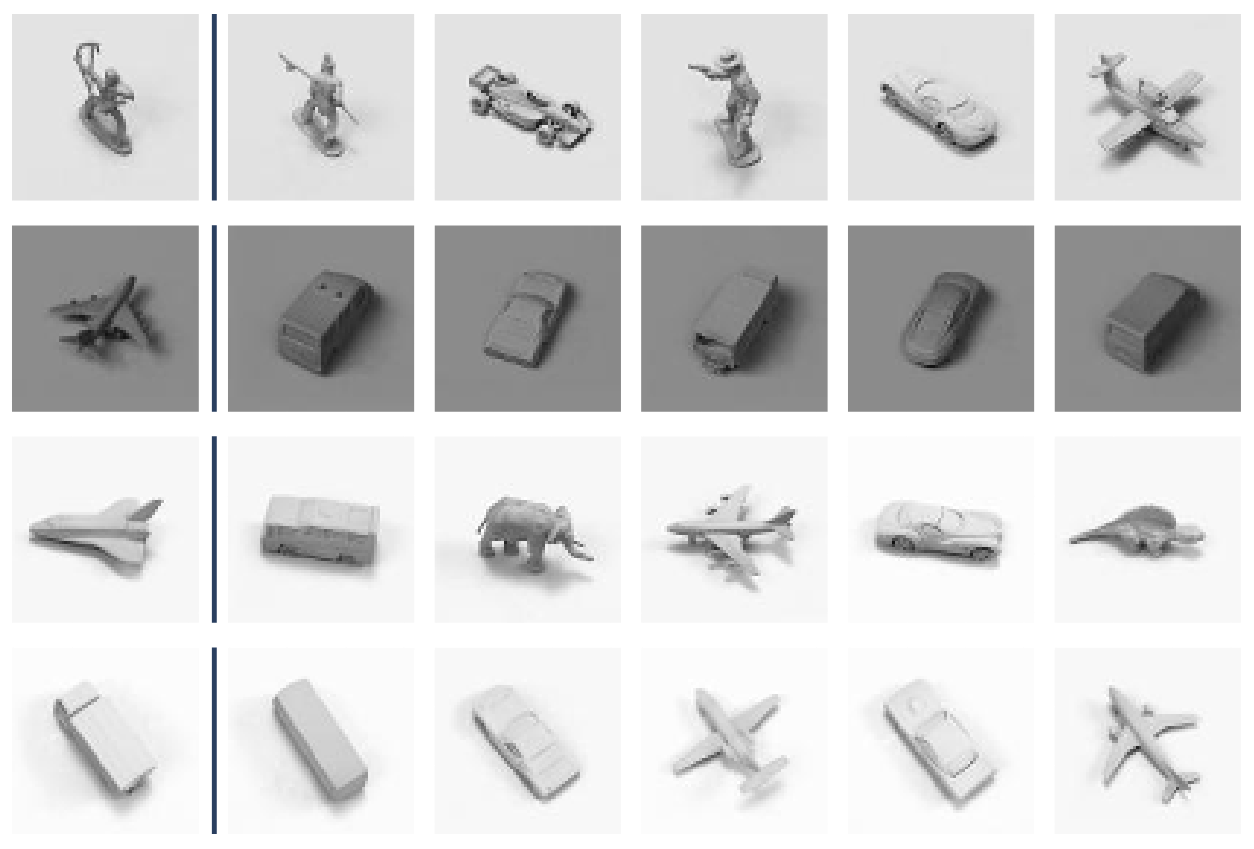} \\ 

\end{tabular}
\caption{Retrieval Examples From SmallNorb.}
\label{fig:retrievals_smallnorb}
\end{figure}

\textbf{Quantitative Analysis.} 
Our numerical retrieval results are presented in Fig.~\ref{tab:retrieval}. Similarly to the earlier probing experiments on Cars3D, LORD achieves the highest retrieval scores among all the baseline methods on this dataset. Our method, DCoDR, convincingly outperforms it, both with and without reconstruction, achieving correct retrieval accuracy of $97\%$. As this is the easiest dataset, DRNET and ML-VAE achieve acceptable results, but underperform DCoDR and LORD due to their lack of invariance. SmallNORB is a much harder task, all baseline methods struggle on this dataset achieving poor retrieval accuracy. We showed in Tab. \ref{tab:disent_and_rep_quality} that these methods have high informativeness and poor invariance on this dataset. This shows that invariance is important for succeeding in cross domain retrieval. DCoDR (with and without reconstruction) is able to retrieve much better matches as it is considerably less biased by the domain. This is backed up by the qualitative analysis of SmallNorb in Fig. \ref{fig:retrievals_smallnorb}. Results on Shapes3D describe a similar case. Although all methods achieve strong informativeness, DCoDR and DCoDR-norec only are able to retrieve perfect matches due to domain invariance. Surprisingly, on this dataset DRNET was able to retrieve strong matches from different domains, despite not being domain invariant at all. Finally, Edges2Shoes showcases a failure of DCoDR-norec. In this case, the augmentations do not provide a strong enough inductive-bias for learning informative representations, making the retrieval considerably weaker than all previous methods. Saying that, when given the inductive-bias of the reconstruction objective, DCoDR exceeds previous methods significantly.

\begin{table}[h!]
    \centering
        \caption{\textbf{\textit{Retrieval Accuracies Comparison.}}}
        \label{tab:retrieval}
        \begin{tabular}{lcccc}
             & Cars3D & SmallNorb & Shapes3D & Edges2Shoes \\
             \midrule
            LORD & 0.92 & 0.08 & 0.71 & 0.75 \\
            DRNET & 0.87 & 0.04 & 0.94 & 0.72 \\
            ML-VAE & 0.71 & 0.03 & 0.48 & 0.78 \\
            \midrule
            DCoDR-norec & \textbf{0.97} & 0.36 & 0.99 & 0.53 \\
            DCODR & \textbf{0.97} & \textbf{0.40} & \textbf{1} & \textbf{0.90} \\
            \bottomrule
    \end{tabular}
\end{table}

\begin{figure}[b!]
\centering
\begin{tabular}{c|c}

DRNET & ML-VAE \\ 
\includegraphics[width=0.4\linewidth]{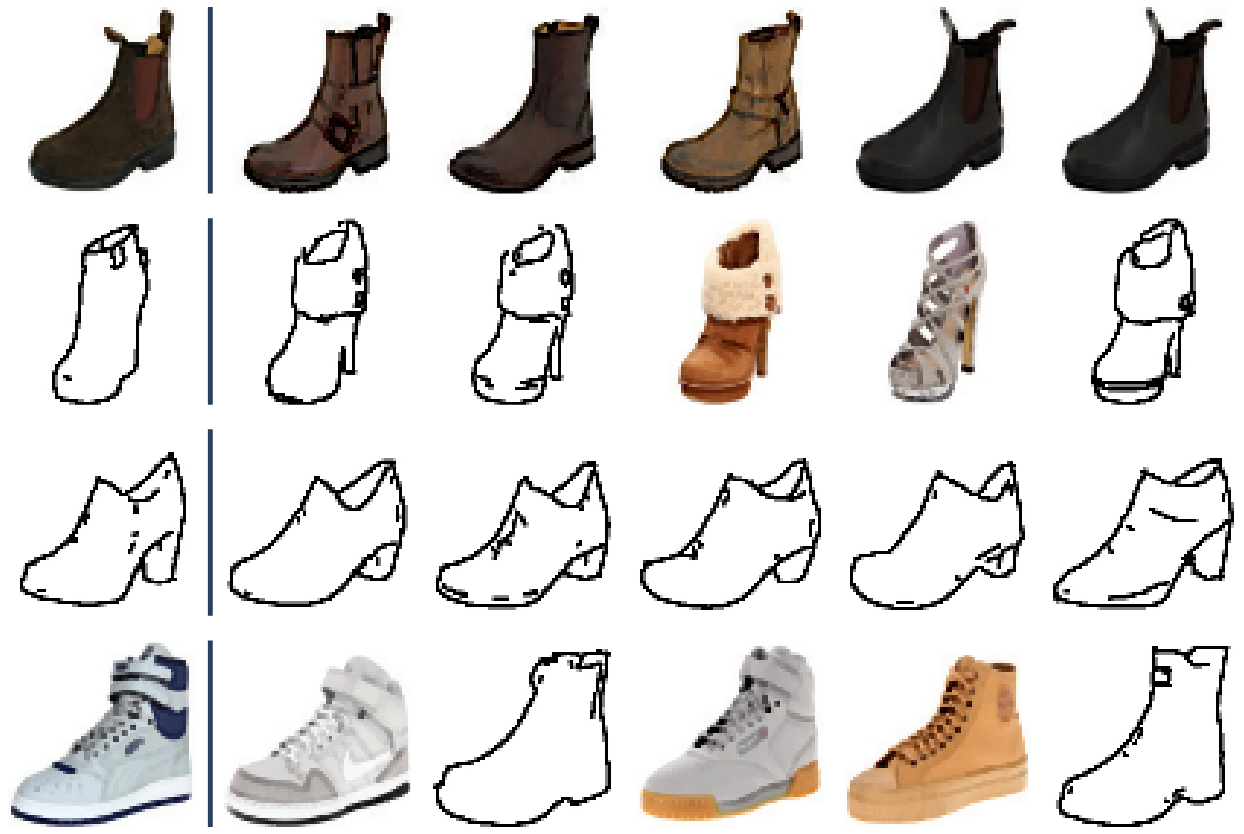} & 
\includegraphics[width=0.4\linewidth]{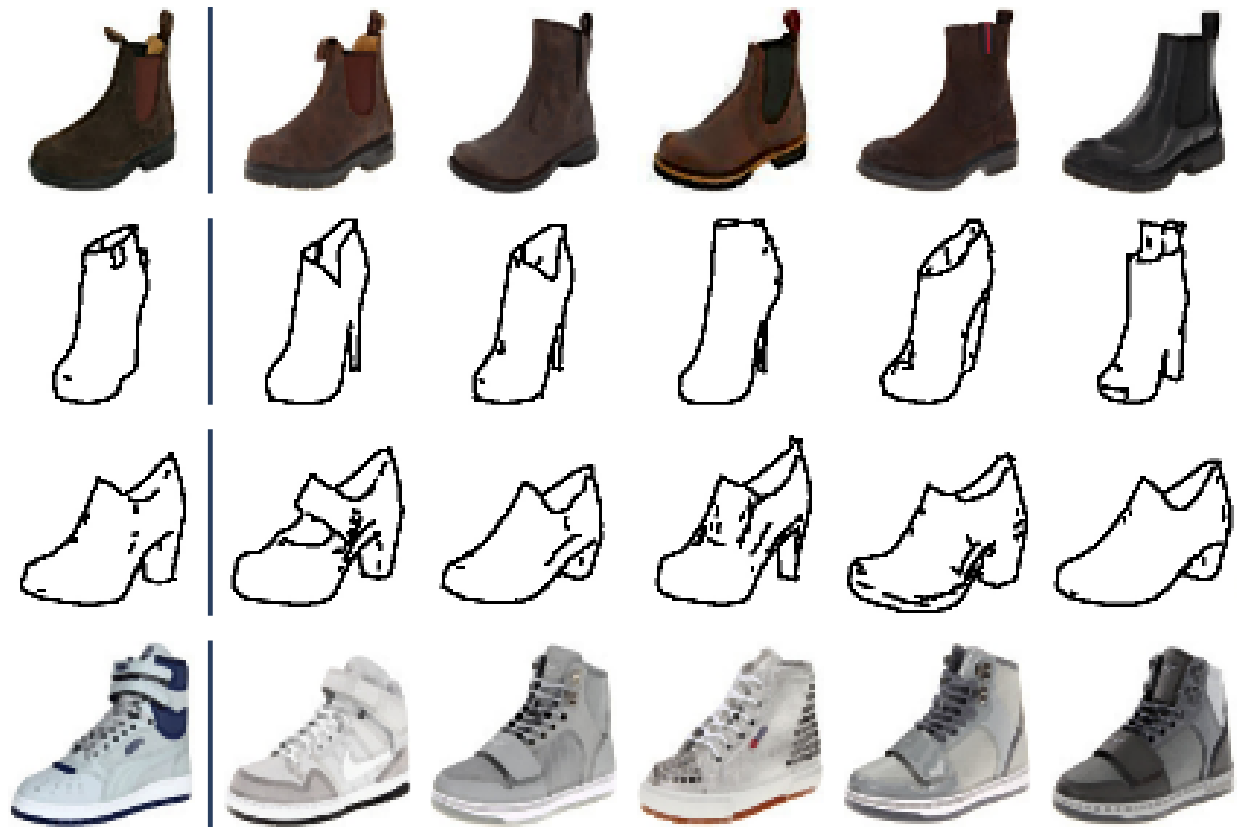} \\
\midrule
 LORD & DCODR  \\
\includegraphics[width=0.4\linewidth]{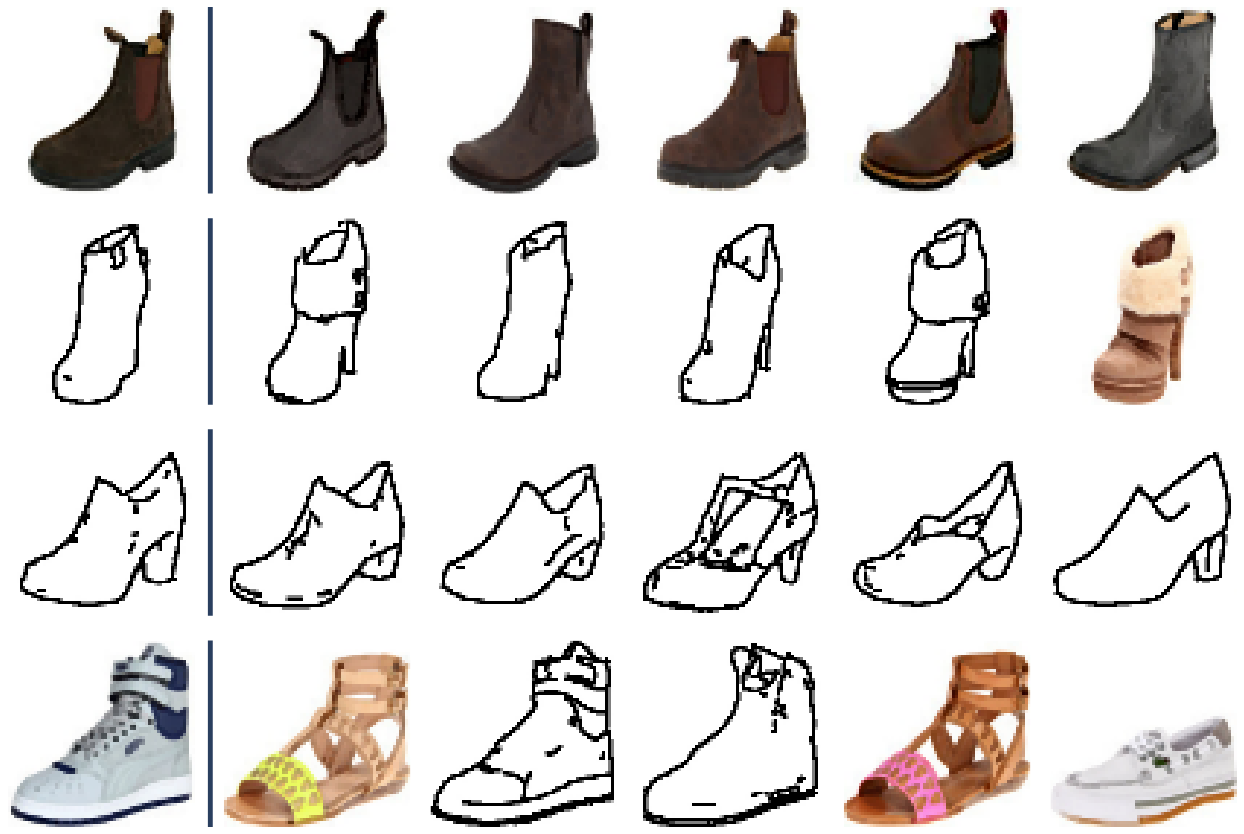}  & 
\includegraphics[width=0.4\linewidth]{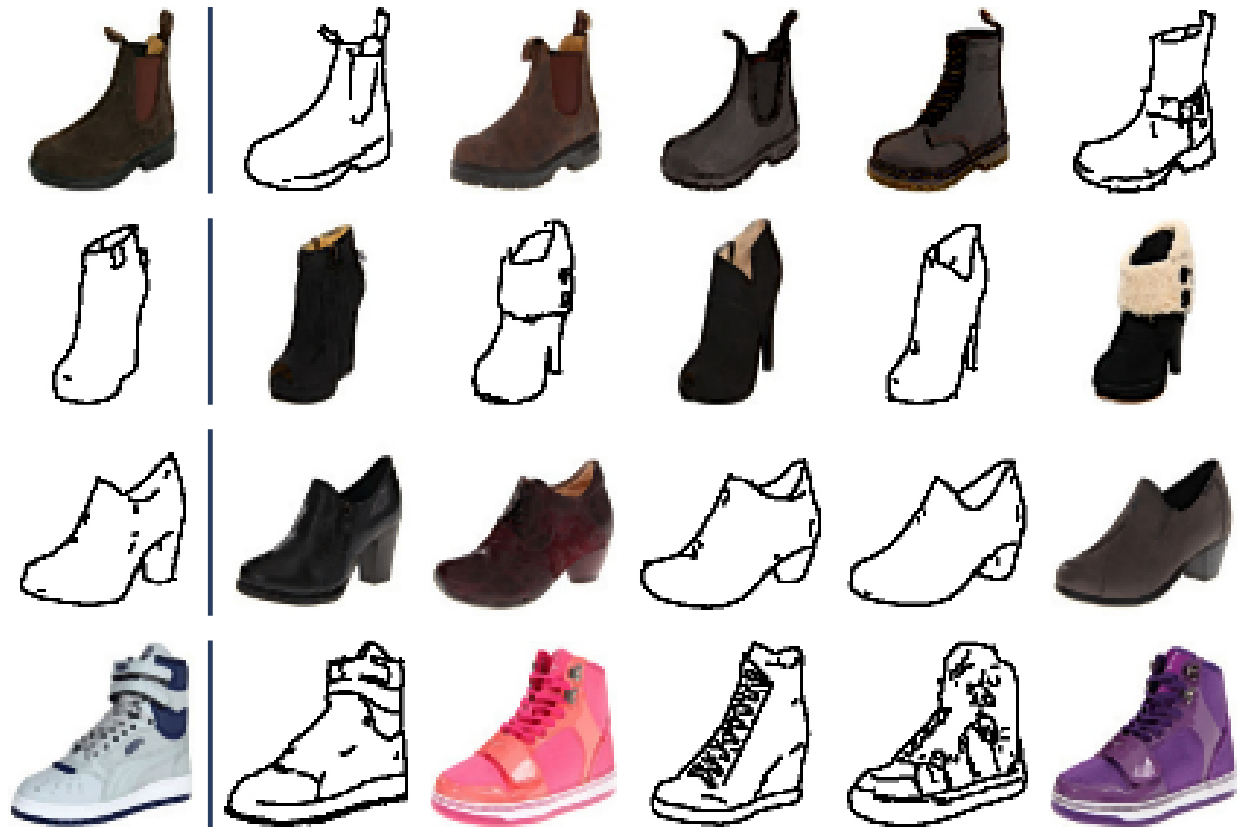}\\ 
\end{tabular}
\caption{Retrieval Examples From Edges2Shoes.}
\label{fig:retrievals_e2s}
\end{figure}

\textbf{Qualitative Analysis.}
We present retrieval results on SmallNorb \cite{smallNORB} and Edges2Shoes \cite{yu2014edges2shoes} datasets in Fig.~\ref{fig:retrievals_smallnorb} and \ref{fig:retrievals_e2s} respectively. We present DCoDR-norec on SmallNorb, and DCoDR on Edges2Shoes (as the reconstruction loss is needed there). On SmallNorb, DRNET and ML-VAE retrieve images from the same domain at the expense of changing the pose, achieving poor retrieval results. While LORD does select images from other domains, the domains are typically similar to the source. DCoDR-norec retrieves images from a variety of domains while preserving the pose. Both LORD and DCoDR-norec struggle with $180^\circ$ flips. For Edges2Shoes, ML-VAE clearly shows lack of domain invariance. DCoDR retrieves more accurate images than DRNET and LORD.

\subsection{Runtime Comparison}
\label{sec:run_time}

We compared our method's runtime with LORD \cite{gabbay2020lord}, the top baseline. All methods were run on a single NVIDIA-RTX6000 for $200$ epochs for all datasets (note that LORD has two stages). Results are presented in Tab. \ref{tab:training_times}. Both DCoDR and DCoDR-norec are faster than LORD. DCoDR-norec  is \textbf{5-10 times faster} than LORD as it does not train a generator nor require perceptual loss computation.

\begin{table}[h!]
\centering
    \centering
        \caption{\textbf{\textit{Training Times {$(\downarrow)$} in Hours.}}}
        \label{tab:training_times}
        \begin{tabular}{lcccc}
             & Cars3D & SmallNorb & Shapes3D (50K) & CelebA \\
            \midrule
            LORD & 7.5 & 15.5 & 18 & 160 \\
            DCODR & 5.5 & 9.5 & 11.5 & 30 \\
            DCoDR-norec & \textbf{1.5} & \textbf{3.5} & \textbf{3.5} & \textbf{9} \\
            \bottomrule
    \end{tabular}
\end{table}

\section{Discussion}

\textbf{The mismatch between conditional reconstruction constraints and informativeness.}  By the data processing inequality, the existence of a deterministic mappings $x = G_{d_{true}}(z)$ (and accordingly $x=E(z)$) implies that $I(y;x|z,d_{true}) = 0$. In other words, all the information about $y$ which exists in $x$, exists in the combination of $z$ and the domain label $d_{true}$ as well. Note this does not imply $I(y;x|z,d') = 0$ for any domain $d'$ but only for the true domain of $x$, $d'=d_{true}$. To be equivalent to Eq.~\ref{eq:completeness_metric}, it was shown by \cite{Johansson2019SupportAI} and \cite{invforda19zhao} that this requires another property from the representations which is \textit{alignment}. Meaning, $p(y|z,d) = p(y|z,d')=p(y|z)$, where $d$ and $d'$ are two different domains and $p(y|z,d)$ is the PDF of $y$'s values given the representation $z$ under domain $d$. Alignment is not guaranteed without additional inductive biases but in practice learned representations are often well aligned.

\noindent \textbf{Inductive bias of generators.} We presented a discriminative variant that, in some cases, competes with the top domain disentanglement methods, which are generative. We believe the reason for the success of conditional generator based methods is two-fold: i) a regularization effect caused by the difficulty of conditional generator training, pushing the representations of different domains to be more aligned. ii) invariance of generators to various image transformations. DCoDR-norec presented \emph{partial} improvements in these two aspects. Pre-trained weights are used for initialization, we hypothesize this acts as a regularizer although not as strong as a conditional generator.
Image augmentations are used, \emph{most of which} are encapsulated in the invariance of generators. To test the invariance of generators to different augmentations, we performed an experiment where we trained autoencoders on several datasets and compared their reconstruction for images with and without augmentations. This motivated our choice of augmentations. For more details, see the App.~\ref{appendix:augmentations_analysis}. Despite DCoDR-norec showing promising results in some cases, we find that all components of our method are needed for sufficient informativeness. We expect that future research will find other augmentations which will result in further improvements.

\section{Limitations} 

(i) \textit{Handling continuous domains.} Our method requires the domain label is discrete, due to our per-domain invariance objective.
    
(ii) \textit{Pre-training.} We showed in Sec. \ref{sec:ablations} that using unsupervised pre-training (MoCo-V2 trained on ImageNet) significantly improves both invariance and informativeness. Although requiring an external dataset is a limitation, we do not believe it is a very serious one for two reasons. Firstly, previous methods, e.g. LORD, often use supervised pre-trained features in their perceptual loss while DCoDR requires unsupervised pre-trained features. Secondly, these weights are available to all, and do not require extra supervision for a new dataset. 
    
(iii) \textit{Image-specific augmentations.} Our method rely on image augmentations, which are not always directly transferable to other modalities such as audio or text. Never-the-less, we believe that augmentations that are helpful to the various metrics can be found in different modalities.

\section{Conclusion}

We presented a new approach for learning domain disentangled representations from images. It uses a per-domain contrastive loss, a reconstruction objective, image augmentations and self-supervised pre-trained encoder initialization. Our method demonstrated results that are better in both invariance and informativeness metrics over the state-of-the-art. 
Our discriminative variant can reach an order of magnitude training speed up than competing state-of-the-art methods.

\section{Acknowledgments}
This work was supported in part by the Federmann Cyber Security Research Center in conjunction with the Israel National Cyber Directorate. Computational resources were kindly supplied by the Oracle for Research program.

\clearpage

\bibliographystyle{unsrt}  
\bibliography{references}  

\clearpage

\appendix

\section{Augmentations Ablation Study}
\label{appendix:augmentations_analysis}

\subsection{Investigating the Inductive Biases of Generative Models}
\label{appendix::gen_inductive_bias}

We designed a principled (although not necessarily optimal) approach for selecting augmentations. The main idea is to select augmentations that have similar invariances as autoencoders - as autoencoders are often used in disentanglement. We conduct an experiment on the CelebA \cite{CelebAMask-HQ}, Cars3D \cite{carsdataset} and Edges2Shoes (shoes only) \cite{yu2014edges2shoes} datasets. The experiment consists of several stages: i) train an autoencoder $AE$ on an image dataset without any augmentations s.t. $\min_{AE} \sum_{x \in {\cal{X}}}\|x - AE(x)\|^2$, where $\cal{X}$ is the training set. ii) transform image from the test set of the same dataset with a range of image augmentations $T$ iii) evaluate the invariance of the outputs of the autoencoder. We use the following two invariance metrics $f_{unnorm},f_{norm}$ for evaluating how much the distance between the original and transformed images change when evaluated on autoencoder outputs. Meaning, for each image augmentation $t \in T$ we measure the followings:
 \begin{equation}
    \label{appendix:lpips_abs}
      f_{unnorm} = dist(AE(x), AE(t(x)))
 \end{equation}
 \begin{equation}
    \label{appendix:lpips_ratio}
      f_{norm} = \frac{dist(AE(x), AE(t(x)))}{dist(x, t(x))}
 \end{equation}
We use a VGG \cite{simonyan2014very} based perceptual loss as the distance function. If an autoencoder is invariant to a particular transformation, \textbf{both} metrics should be small. The normalized metric is sensitive to smaller transformations, and the unnormalized metric is sensitive to larger transformations. We average the results over different datasets in Tab. \ref{appendix:inductive_bias_gen_ave} while the full results are shown in Tab. \ref{appendix:tab:inductive_bias_gen_full}.

\begin{table}[hbt!]
    \centering
        \caption{An evaluation of the invariance of autoencoders to different image transformations}
        \label{appendix:inductive_bias_gen_ave}
        \begin{tabular}{lcc}
            & \multicolumn{2}{c} {\textbf{Average}} \\
            \cmidrule(lr){2-3}
             & \textbf{$f_{norm}$} & \textbf{$f_{unnorm}$} \\
            \midrule
            Horizontal Flip & 0.868 & 0.2489 \\
            Vertical Flip & 0.791 & 0.3125 \\
            Low Contrast & 1.841 & 0.0903 \\
            Low Brightness & 0.759 & 0.1666 \\
            Color Rotation & 0.876 & 0.1020 \\
            Random Erase \cite{random_erasing} & 0.796 & 0.1967 \\
            Affine Transformation & \textbf{0.650} & 0.382 \\
            GrayScale & 0.787 & \textbf{0.0570} \\
            Crop & 0.842 & 0.2053 \\
            High Brightness & 0.806 & \textbf{0.0759} \\
            \textbf{High Contrast} & \textbf{0.433} & \textbf{0.0572} \\
            \textbf{High Saturation} & \textbf{0.579} & \textbf{0.0261} \\
            \textbf{Gaussian Blurring} & \textbf{0.315} & \textbf{0.0378} \\
            \bottomrule
    \end{tabular}
\end{table}

We observe that autoencoders are highly invariant to gaussian blurring, high saturation and high contrast. As these are the inductive biases of generative methods, it suggests that providing these biases to discriminative methods can potentially transfer some of the attractive qualities of generative methods.

\subsection{Additional Experiments}
\label{appendix:sec:aug_additional_exps}

\begin{figure}[h!]
\centering
\begin{tabular}{c}
Invariance \\ 
\includegraphics[width=0.9\linewidth]{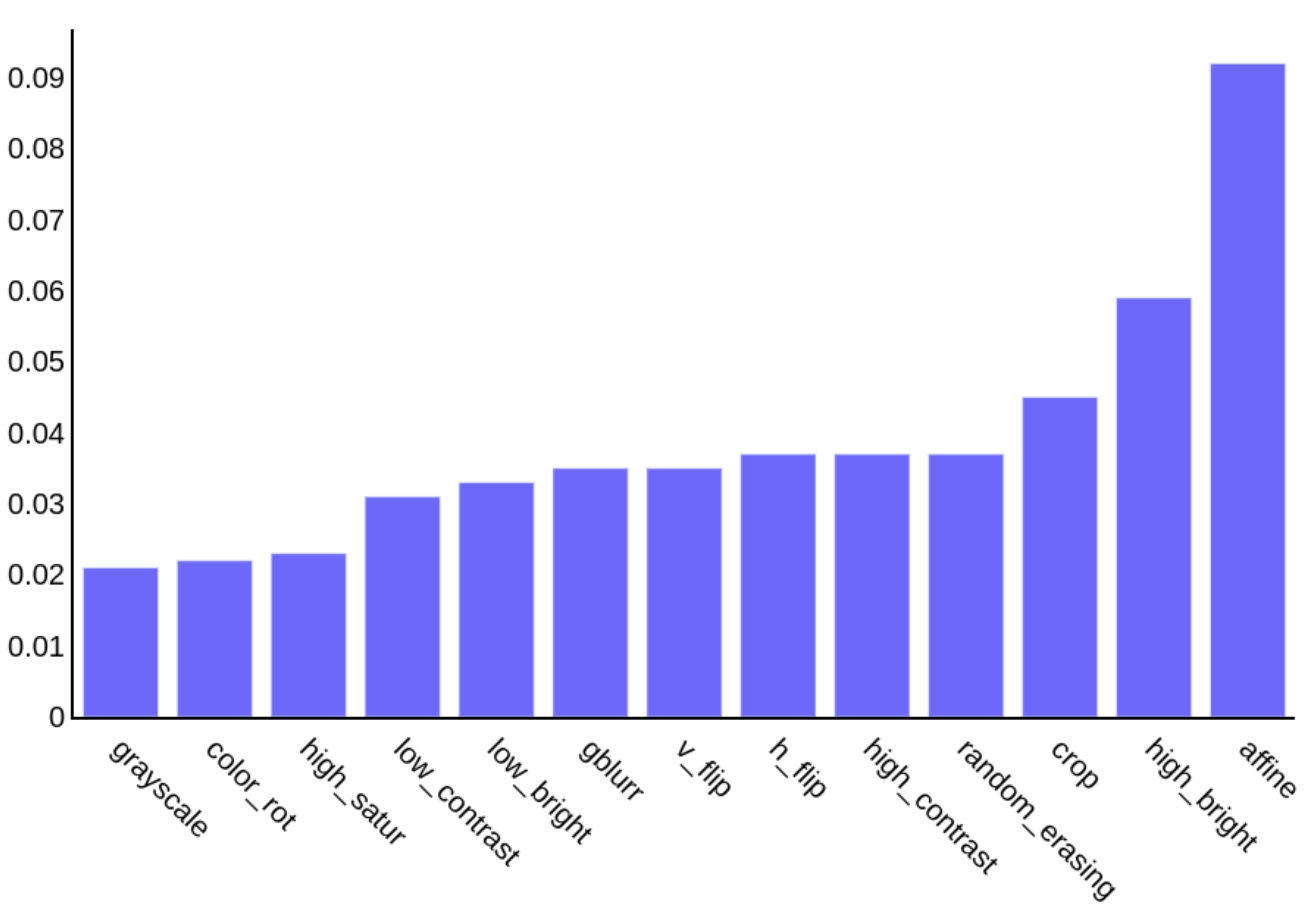} \\
\\
Informativeness \\
\includegraphics[width=0.9\linewidth]{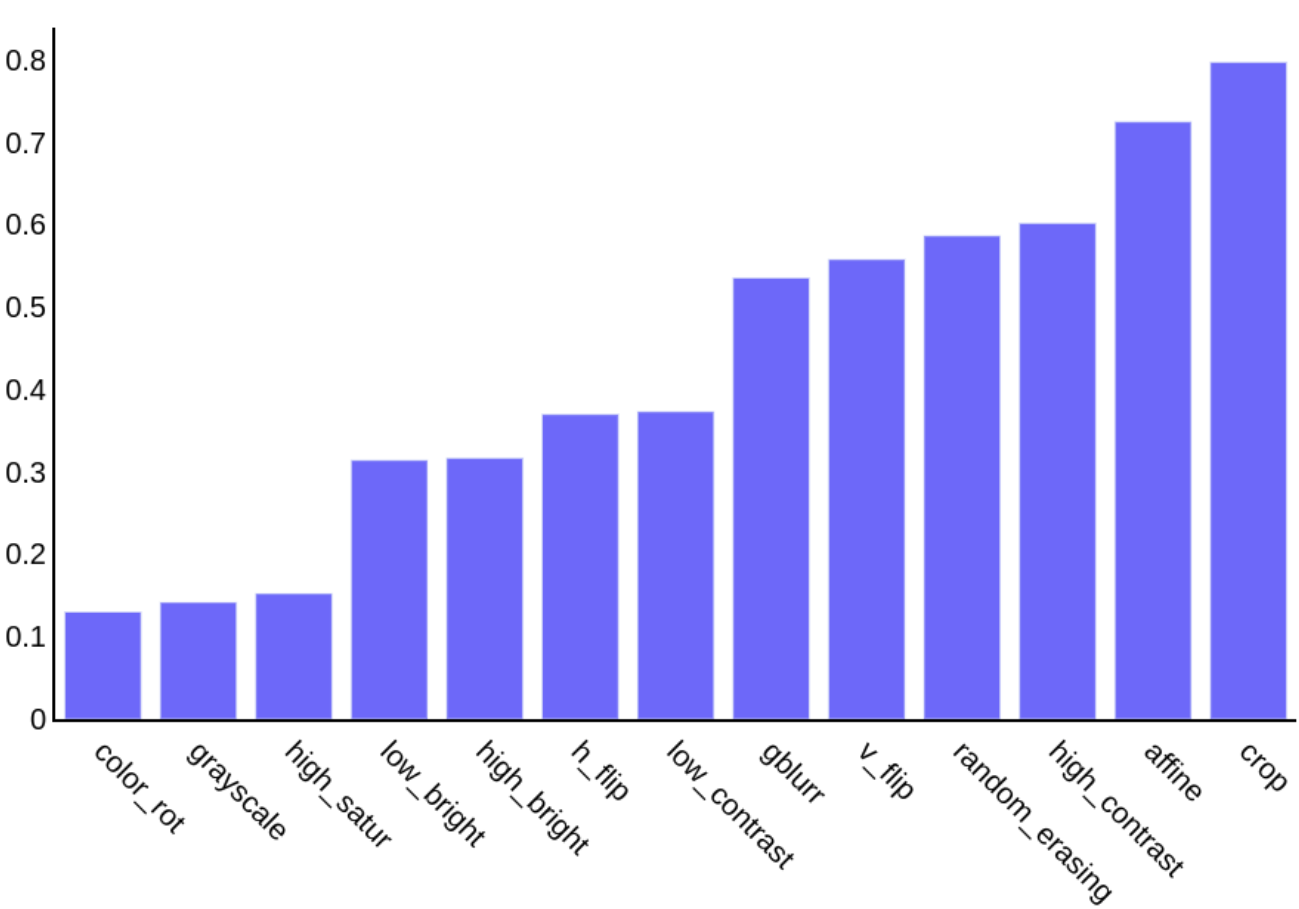} \\
\end{tabular}
\caption{Invariance and Informativeness of different augmentations.}
\label{appendix:fig::aug_analysis}
\end{figure}

In addition to the experiments presented in Sec. \ref{appendix::gen_inductive_bias}, we perform another analysis to look for other augmentations that might be useful for our method. We propose the following experiment: training DCoDR-norec with only a single augmentation. We measure the invariance and informativeness resulted by this augmentation, mostly considering the \emph{invariance}, in order to not interfere with the disentanglement process. We show our results over the SmallNorb \cite{smallNORB} dataset in Fig. \ref{appendix:fig::aug_analysis}. Our results show that the crop augmentation performed well on invariance (as we defined it in Sec.~\ref{sec:crit}), and extremely well for the informativeness. For those reasons we decided to insert it to our augmentations set as well, reaching a total of 4: i) Gaussian Blurring ii) High Contrast iii) High Saturation iv) Cropping. Note, that in specific cases (e.g. Edges2Shoes) adding or removing augmentations might result in even better metrics, and we encourage future research to find a better selection method of a set of augmentations, which might be specific for each dataset. Saying that, our choice of transformations is reasonable, and not particularly optimized to a specific dataset.

\FloatBarrier

\FloatBarrier

\subsection{Augmentations Detailed Experimental Results}
 
We show the results over all augmentations in each dataset. We measure the autoencoder's invariance to several augmentations (not to be confused with our invariance metric from other sections) by $f_{unnorm}$ (\ref{appendix:lpips_abs}) and $f_{norm}$ (\ref{appendix:lpips_ratio}). Results are shown in Tab. \ref{appendix:tab:inductive_bias_gen_full}. We present the details of each augmentation in PyTorch TorchVision library \cite{pytorch} -style notations (but note this is psuedo-code):
 
 \begin{itemize}
     
     \item Horizontal Flipping: RandomHorizontalFlip(p=1.)
     
     \item Vertical Flipping: RandomVerticalFlip(p=1.)
     
     \item Low Contrast: ColorJitter(contrast=(0.3, 0.8))
     
     \item Low Brightness: ColorJitter(brightness=(0.3, 0.8))
     
     \item Color Rotation: ColorJitter(hue=(0.2, 0.5) or hue=(-0.5, -0.2))
     
     \item Random Erase: RandomErasing(p=1.)
     
     \item Affine Transformation: RandomAffine(degrees=(-40, 40), fill=255)
     
     \item GrayScale: RandomGrayscale(p=1.)
     
     \item Crop: RandomResizedCrop(scale=(0.5, 1.))
     
     \item High Brightness: ColorJitter(brightness=(1.4, 1.8))
     
     \item High Contrast: ColorJitter(contrast=(1.8, 3.0))
     
     \item High Saturation: ColorJitter(saturation=(1.8, 3.0))
     
     \item Gaussian Blurring: GaussianBlur(kernel\_size=5, sigma=1.)

 \end{itemize}
 
\begin{table}[h!]
    \caption{A detailed evaluation of the invariance of autoencoders to different image transformations. $f_{unnorm}$ (\ref{appendix:lpips_abs}) and $f_{norm}$ (\ref{appendix:lpips_ratio}) are described in Sec.~\ref{appendix::gen_inductive_bias}}
    \label{appendix:tab:inductive_bias_gen_full}
    \centering
        \begin{tabular}{lcccccc}
            & \multicolumn{2}{c}{\textbf{Cars3D}} & \multicolumn{2}{c}{\textbf{Edges2Shoes}} & \multicolumn{2}{c}{\textbf{CelebA}} \\ \cmidrule(lr){2-3} \cmidrule(lr){4-5} \cmidrule(lr){6-7}
             & $f_{norm}$ & $f_{unnorm}$ & $f_{norm}$ & $f_{unnorm}$ & $f_{norm}$ & $f_{unnorm}$ \\
            \midrule
            Horizontal Flip & 0.861 & 0.1212 & 0.868 & 0.3387 & 0.876 & 0.2869 \\
            Vertical Flip & 0.853 & 0.1991 & 0.770 & 0.3458 & 0.751 & 0.3925 \\
            Low Contrast & 1.231 & 0.0710 & 3.537 & 0.1442 & 0.756 & 0.0557 \\
            Low Brightness & 0.749 & 0.1978 & 1.098 & 0.2265 & 0.429 & 0.0755 \\
            Color Rotation & 1.118 & 0.0725 & 1.015 & 0.0829 & 0.496 & 0.1507 \\
            Random Erase \cite{random_erasing} & 0.728 & 0.1994 & 0.884 & 0.1994 & 0.776 & 0.1968 \\
            Affine Transformation & 0.479 & 0.2658 & 0.734 & 0.4116 & 0.736 & 0.4675 \\
            GrayScale & 0.957 & 0.0479 & 0.939 & 0.0502 & 0.464 & 0.0729 \\
            Crop & 0.808 & 0.1683 & 0.830 & 0.2096 & 0.889 & 0.2380 \\
            High Brightness & 0.355 & 0.0372 & 1.188 & 0.0956 & 0.875 & 0.0950 \\
            \textbf{High Contrast} & 0.312 & 0.0356 & 0.454 & 0.0499 & 0.534 & 0.0862 \\
            \textbf{High Saturation} & 0.337 & 0.0095 & 0.768 & 0.0243 & 0.632 & 0.0445 \\
            \textbf{Gaussian Blurring} & 0.080 & 0.0111 & 0.180 & 0.0196 & 0.055 & 0.0071 \\
            \bottomrule
    \end{tabular}
\end{table}

\subsection{Other Datasets}

We leave Shapes3D out of this ablation study in order to validate that our selected augmentations are able to transfer to unseen datasets. We hypothesize our conservative selection process has a higher chance to better transfer to other datasets. That being said, and as noted in Sec.~\ref{appendix:sec:aug_additional_exps}, this augmentation selection can be significantly improved, which we leave for future work.

\section{Implementation Details}

In this section, we elaborate the implementation details of our algorithm to ensure reproducibility. Note that code is also included.

\textbf{Datasets} We use 5 datasets for our evaluation: Cars3D \cite{carsdataset}, SmallNorb \cite{smallNORB}, Shapes3D \cite{3dshapes18}, CelebA \cite{CelebAMask-HQ} and Edges2Shoes \cite{yu2014edges2shoes}. All datasets are used in 64x64 resolution. Shapes3D is subsampled as described in Sec.~\ref{sec:exp}. Since our method uses contrastive learning over each domain separately, it requires several examples from each class to achieve uniformity. For this reason, and for the CelebA dataset alone, we limit the minimum number of samples in each class to be 20, ignoring all classes which have 19 or less samples in the training set alone. Note that i) we do that for our method alone, while the other methods use the entire training dataset ii) the test set is unchanged, except for classes with a single example which are removed, as they cannot be evaluated using a classifier. This filtering causes our method to train on $62\%$ of the training classes which are $81\%$ of the training samples in CelebA.

\textbf{Architecture.} We use a ResNet50 encoder. For SmallNORB \cite{smallNORB} and Shapes3D \cite{3dshapes18} we add 3 fully-connected layers at the end of the encoder. In line with other methods such as LORD, for our generator we use a VGG based perceptual loss pre-trained on ImageNet.

\textbf{Optimization hyperparameters.} We use a learning rate of $0.0001$ for the encoder and $0.0003$ for the generator, except for CelebA where we use $0.001$ for both. We train our method for $200$ epochs, using a batch size of 128, composed from 32 images drawn from 4 different classes.

\textbf{Temperature.} We use $0.2$ for Cars3D, SmallNorb and CelebA. For Shapes3D and Edges2Shoes we use $0.1$.

\textbf{Reconstruction Loss Weight.} We use a scalar constant to weight the importance of the reconstruction loss relative to the contrastive loss. We use $0.3$ for all datasets.

\section{Complete Experimental Results}
\label{appendix:sec:complete_results}

\subsection{Representation Evaluation}

We present the complete results of all experiments on the SmallNorb \cite{smallNORB} and Shapes3D \cite{3dshapes18} datasets in Tab.  \ref{appendix:tab::full_res_norb} and Tab.~\ref{appendix:tab::full_res_shapes} accordingly. For Cars3D \cite{carsdataset}, CelebA \cite{CelebAMask-HQ} we only predict a single attribute (full pose - azimuth and elevation, landmarks regression) therefore the full results have already been presented in Sec.~\ref{subsec:eval_rep}.

\begin{table}[h!]
\centering
    \caption{Representation evaluation for each factor of SmallNorb.}
    \label{appendix:tab::full_res_norb}
    \begin{tabular}{lcccc}
        & \textbf{Domain} & \textbf{Azimuth} & \textbf{Elevation} & \textbf{Lighting} \\ 
         \midrule
        LORD & 0.393 & 0.731 & 0.384 & 0.895 \\
        DrNet & 0.953 & 0.973 & 0.766 & 0.957 \\
        ML-VAE & 0.968 & 0.982 & 0.868 & 0.982 \\
        \midrule
        DCoDR-norec & 0.071 & 0.619 & 0.594 & 0.977 \\
        DCoDR & 0.143 & 0.684 & 0.695 & 0.977 \\
        \midrule
        Optimal & 0.021 & 1 & 1 & 1 \\
        \bottomrule
\end{tabular}
\end{table}

\begin{table}[h!]
    \centering
    \caption{Representation evaluation for each factor of Shapes3D.}
    \label{appendix:tab::full_res_shapes}
    \begin{tabular}{lcccccc}
    & & \multicolumn{3}{c}{\emph{Colors}} \\
    \cmidrule(lr){3-5} & \textbf{Domain} & \textbf{Floor} & \textbf{Wall } & \textbf{Object}  & \textbf{Scale} & \textbf{Orientation} \\
     \midrule
    LORD & 0.703 & 0.998 & 0.999 & 0.990 & 0.991 & 0.999 \\
    DrNet & 0.892 & 1 & 1 & 1 & 0.999 & 1 \\
    ML-VAE & 0.999 & 1 & 1 & 1 & 1 & 1 \\
    \midrule
    DCoDR-norec & 0.246 & 1 & 0.999 & 0.998 & 0.991 & 0.997 \\
    DCoDR & 0.245 & 0.999 & 0.999 & 0.999 & 1 & 0.999 \\
    \midrule
    Optimal & 0.25 & 1 & 1 & 1 & 1 & 1 \\
    \bottomrule
\end{tabular}

\end{table}

\FloatBarrier

\subsection{Retrieval}

\textbf{Complete Retrieval Results.} We display the complete results of the retrieval task, showing retrieval accuracies for both each factor separately and perfect match retrievals. For Cars3d we have only a single attribute (pose) meaning results are displayed in Tab.~\ref{tab:retrieval}. Additional results are presented in Tab.~\ref{appendix:tab:retrieve_norb}, \ref{appendix:tab:retrieve_shapes3d} and \ref{appendix:tab:retrieve_e2s}.

\textbf{Error Margin in the Retrieval Task.} As the changes in the azimuth property of the SmallNorb dataset are relatively small, we decided to allow an error margin of $3$ for this attribute. For all other attributes in all the other datasets we require a perfect match.

\begin{table}[h!]
    \centering
        \caption{SmallNorb retrieval accuracies.}
        \label{appendix:tab:retrieve_norb}
        \begin{tabular}{lcccc}
            & \textbf{Azimuth} & \textbf{Elevation} & \textbf{Lighting} & \textbf{All} \\
            \midrule
            LORD & 0.90 & 0.30 & 0.79 & 0.08 \\
            DrNet & 0.99 & 0.35 & 0.67 & 0.04 \\
            ML-VAE & 0.99 & 0.56 & 0.43 & 0.026 \\
            \midrule
            DCoDR-norec & 0.65 & 0.52 & 0.97 & 0.36 \\
            DCoDR & 0.65 & 0.61 & 0.97 & 0.40 \\
            \bottomrule
    \end{tabular}
\end{table}

\begin{table}[h!]
    \centering
        \caption{Shapes3D retrieval accuracies.}
        \label{appendix:tab:retrieve_shapes3d}
        \begin{tabular}{lcccccc}
            & \multicolumn{3}{c}{\emph{Colors}} \\
            \cmidrule(lr){2-4} & \textbf{Floor} & \textbf{Wall} & \textbf{Object} & \textbf{Scale} & \textbf{Orientation} & \textbf{All} \\
            \midrule
            LORD & 0.99 & 1 & 0.94 & 0.96 & 0.81 & 0.71 \\
            DrNet & 1 & 0.99 & 1 & 0.96 & 0.99 & 0.94 \\
            ML-VAE & 0.71 & 0.98 & 0.79 & 0.99 & 1 & 0.48 \\
            \midrule
            DCoDR-norec & 1 & 1 & 1 & 1 & 1 & 0.99 \\
            DCoDR & 1 & 1 & 1 & 1 & 1 & 1 \\
            \bottomrule
    \end{tabular}
\end{table}

\begin{table}[h!]
    \centering
        \caption{Edges2Shoes retrieval accuracies.}
        \label{appendix:tab:retrieve_e2s}
        \begin{tabular}{lccc}
            & \textbf{Shoe} \textbf{Type} & \textbf{Gender} & \textbf{All} \\
            \midrule
            LORD & 0.94 & 0.79 & 0.75 \\
            DrNet & 0.93 & 0.76 & 0.72 \\
            ML-VAE & 0.96 & 0.81 & 0.78 \\
            \midrule
            DCoDR-norec & 0.79 & 0.62 & 0.53 \\
            DCoDR & 0.98 & 0.91 & 0.90 \\
            \bottomrule
    \end{tabular}
\end{table}

\end{document}